\pdfoutput=1

\documentclass[11pt]{article}

\usepackage{acl}

\usepackage{times}
\usepackage{latexsym}

\usepackage[T1]{fontenc}

\usepackage[utf8]{inputenc}

\usepackage{microtype}
\usepackage{graphicx}
\usepackage{hyperref}
\usepackage{booktabs}
\usepackage{amssymb}
\usepackage{pifont}
\newcommand{\cmark}{\ding{51}}%
\newcommand{\xmark}{\ding{55}}%
\usepackage{cascadia-code}
\usepackage[colorinlistoftodos,prependcaption]{todonotes}

\newcommand\blfootnote[1]{%
 \begingroup
  \renewcommand\thefootnote{}\footnote{#1}%
   \addtocounter{footnote}{-1}%
 \endgroup
}

\title{Source Code Data Augmentation for Deep Learning: A Survey}

\author{Terry Yue Zhuo$^{1,2\dagger}$, Zhou Yang$^3$, Zhensu Sun$^3$, \\ \textbf{Yufei Wang}$^4$,  \textbf{Li Li}$^5$, \textbf{Xiaoning Du}$^1$, \textbf{Zhenchang Xing}$^{2,6}$, \textbf{David Lo}$^3$\\
        $^1$ Monash University
        $^2$ CSIRO's Data61 
        $^3$ Singapore Management University\\
        $^4$ Huawei Noah's Ark Lab 
        $^5$ Beihang University
        $^6$ Australian National University
        \\
       \texttt{terry.zhuo@monash.edu}
        }

\begin{document}
\maketitle

\begin{abstract}
The increasingly popular adoption of deep learning models in many critical source code tasks motivates the development of data augmentation (DA) techniques to enhance training data and improve various capabilities (e.g., robustness and generalizability) of these models.
Although a series of DA methods have been proposed and tailored for source code models, there lacks a comprehensive survey and examination to understand their effectiveness and implications. This paper fills this gap by conducting a comprehensive and integrative survey of data augmentation for source code, wherein we systematically compile and encapsulate existing literature to provide a comprehensive overview of the field.
We start with an introduction of data augmentation in source code and then provide a discussion on  major representative
approaches.
Next, we highlight the general strategies and techniques to optimize the DA quality.
Subsequently, we underscore techniques useful in real-world source code scenarios and downstream tasks. 
Finally, we outline the prevailing challenges and potential opportunities for future research. 
In essence, we aim to demystify the corpus of existing literature on source code DA for 
 deep learning, and foster further exploration in this sphere. Complementing this, we present a continually updated GitHub repository that hosts a list of update-to-date papers on DA for source code modeling, accessible at \url{https://github.com/terryyz/DataAug4Code}.
\end{abstract}

\blfootnote{\textbf{$\dagger$} Corresponding author.}
\begin{figure*}[htbp]
  \centering
  \begin{minipage}[b]{0.42\textwidth}
    \includegraphics[width=\textwidth]{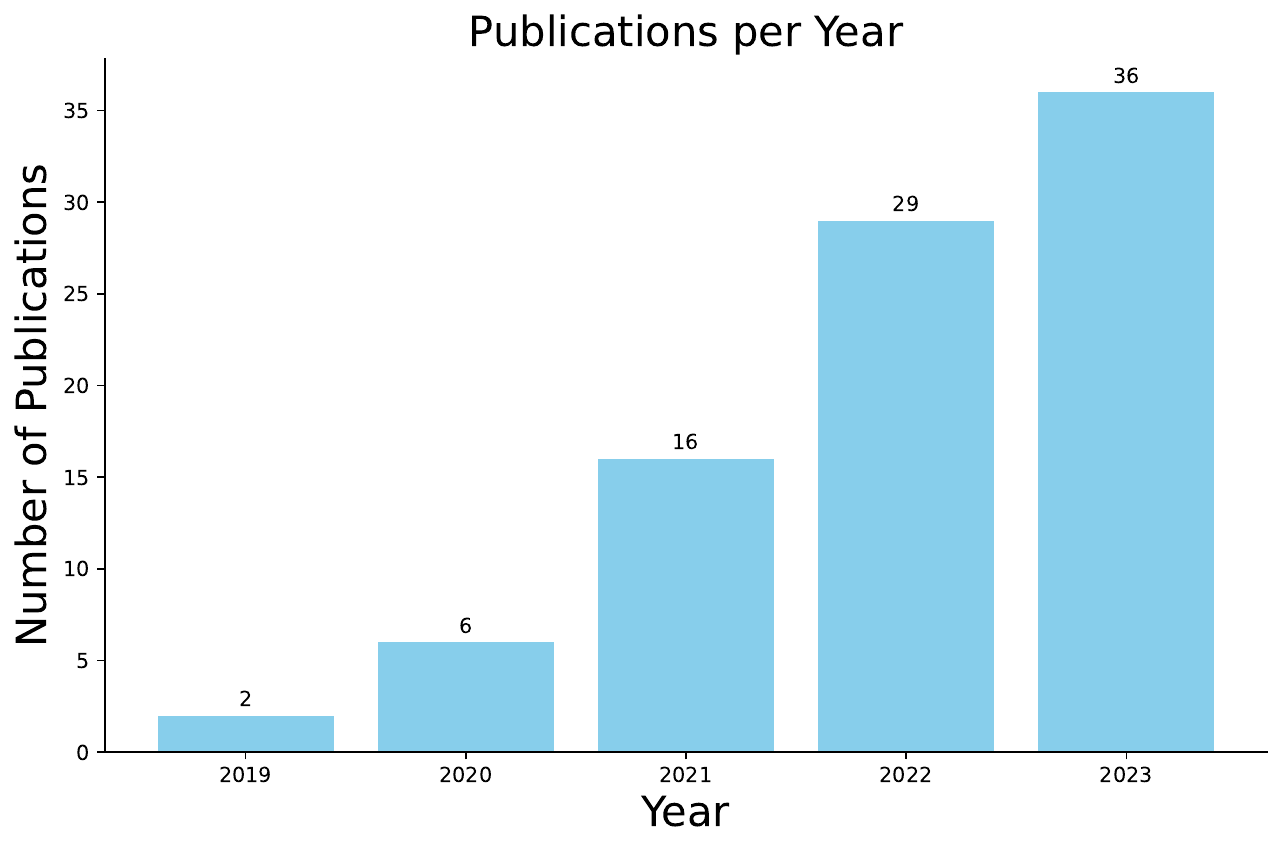}
    \caption{Yearly publications on the topic of ``Source Code DA for Deep Learning''. Data Statistics as of November 2023.}
    \label{fig:pub_year}
  \end{minipage}
  \hfill
  \begin{minipage}[b]{0.42\textwidth}
    \includegraphics[width=\textwidth]{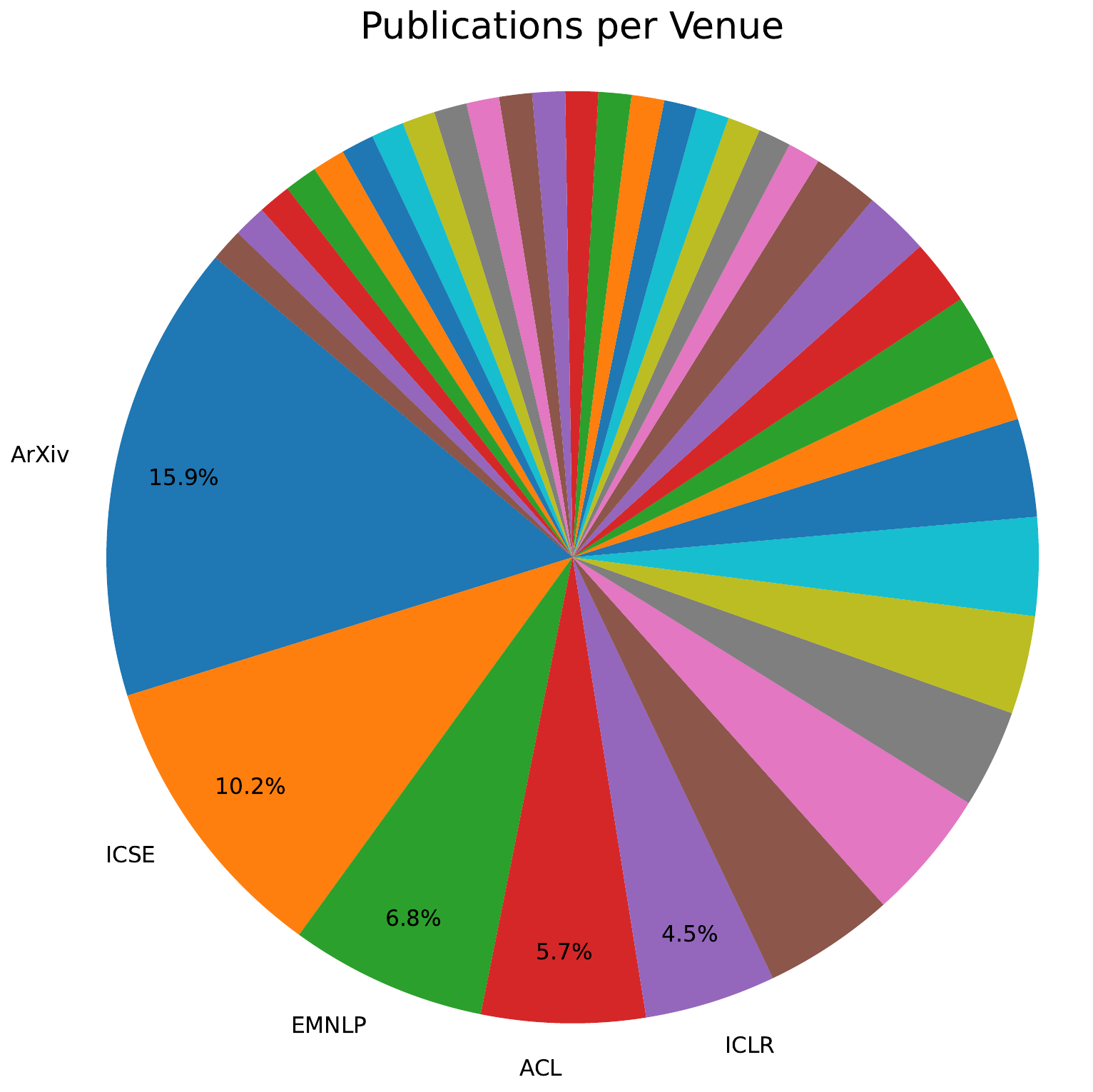}
    \caption{Venue Distribution of the collected publications.}
    \label{fig:pub_venue}
  \end{minipage}
\end{figure*}
\section{Introduction} 
Data augmentation (DA) is a technique used to increase the variety of training examples without collecting new data. 
It has gained popularity in recent machine learning (ML) research, with methods like back-translation~\cite{sennrich2015improving,shiri2022paraphrasing}, Mixup~\cite{zhang2018mixup}, and synthetic audio~\cite{9609154} being widely adopted in natural language processing (NLP), computer vision (CV), and speech recognition. 
These techniques have significantly improved the performance of data-centric models in low-resource domains. For example, \citet{fadaee2017data} obtain substantial improvements for low-resource machine translation via DA, where the translation system is trained with the bilingual pairs synthesized from a limited training corpus.

However, DA has not yet been fully explored in source code modeling, which is the intersection of ML and software engineering (SE). Source code modeling is an emerging area that applies ML techniques to solve various source code tasks such as code completion~\cite{yin2017syntactic}, code summarization~\cite{mcburney2014automatic}, and defect detection~\cite{wang2016automatically}, by training models on a vast amount of data available in open-source repositories~\cite{Allamanis2017ASO}. Source code data typically has two modalities: the programming language (e.g., Python and Java) and the natural language (e.g., doc-strings and code comments), which complement each other. 
Such dual-modality nature of source code data presents unique challenges in tailoring DA for NLP to source code models. For example, the context of a sentence can be relatively standalone or derived from a few surrounding sentences in many NLP tasks~\cite{feng2021survey}. However, in source code, the context can span across multiple functions or even different files, due to the widespread use of function calls, object-oriented programming, and modular design. Therefore, we argue that DA methods for source code would need to take this extended context into account, to avoid introducing errors or changing the original program's behavior.
In addition, source code follows strict syntactic rules that are specified using context-free grammar. 
Consequently, conventional NLP data augmentation methods, such as token substitution with similar words, may make the augmented source code fail to compile and introduce erroneous knowledge for training models.

Despite such challenges, there has been increasing interest and demand for DA for source code modeling. With the growing accessibility of large, off-the-shelf, pre-trained source code models via learning from large-scale corpora~\cite{Chen2021EvaluatingLL, Li2023StarCoderMT,Allal2023SantaCoderDR}, there is a growing focus on applying these models to real-world software development. For instance, \citep{husain2019codesearchnet} observed that many programming languages are low-resource, emphasizing the importance of DA to improve model performance and robustness on unseen data. 


This study aims to bring attention from both ML and SE communities to this emerging field.
As depicted in Figure~\ref{fig:pub_year}, the relevant publications have been increasing in the recent five years.
More precisely, we have compiled a list of 89 core papers from the past five years, mainly from premier conferences and journals in both the ML and SE disciplines as shown in Figure~\ref{fig:pub_venue} (with 62 out of 89 papers published in Core Rank A/A* venues\footnote{We refer to the venues listed at \url{http://portal.core.edu.au/conf-ranks/} and \url{http://portal.core.edu.au/jnl-ranks/}.}). Given the escalating interest and burgeoning research in this domain, it is timely for our survey to (1) provide a comprehensive overview of DA for source code models, and (2) pinpoint key challenges and opportunities to stimulate and guide further exploration in this emerging field. To the best of our awareness, our paper constitutes the first comprehensive survey offering an in-depth examination of DA techniques for source code models.

The structure of this paper is organized as follows:
\begin{itemize}
\item Section~\ref{sec:da} offers a thorough review of three categories of DA for source code modeling: rule-based (\ref{subsec:rule}), model-based (\ref{subsec:model}), and example interpolation-based (\ref{subsec:ei}) techniques.
\item Section~\ref{sec:tricks} provides a summary of prevalent strategies and techniques designed to enhance the quality of augmented data, encompassing method stacking (\ref{subsec:stack}) and optimization (\ref{subsec:optimize}).
\item Section~\ref{sec:scenario} articulates various beneficial source code scenarios for DA, including adversarial examples for robustness (\ref{subsec:adv}), low-resource domains (\ref{subsec:low}), retrieval augmentation (\ref{subsec:retrievalaug}), and contrastive learning (\ref{subsec:contrastive}).

\item Section~\ref{sec:task} delineates DA methodologies for common source code tasks, such as code authorship attribution (\ref{subsec:author}), clone detection (\ref{subsec:clone}), defect detection and repair (\ref{subsec:defect}), code summarization (\ref{subsec:code_summ}), code search (\ref{subsec:search}), code completion (\ref{subsec:completion}), code translation (\ref{subsec:translation}), code question answering (\ref{subsec:cqa}), problem classification (\ref{subsec:problem_class}), method name prediction (\ref{subsec:name_pred}), and type prediction (\ref{subsec:type_pred}).

\item Section~\ref{sec:challenge} expounds on the challenges and future prospects in the realm of DA for source code modeling.
\end{itemize}

Through this work, we hope to emulate prior surveys which have analyzed DA techniques for other data types, such as text~\cite{feng2021survey}, time series~\cite{Wen2020TimeSD}, and images~\cite{Shorten2019ASO}. 
Our intention is to pique further interest, spark curiosity, and encourage further research in the field of data augmentation, specifically focusing on its application to source code.

\section{Background}
\subsection{What are source code models?} Source code models are trained on large-scale corpora of source code and therefore able to model the contextual representations of given code snippets~\cite{Allamanis2017ASO}. In the early stage, researchers have attempted to leverage deep learning architectures like LSTM~\cite{gu2016deep} and Seq2Seq~\cite{yin2017syntactic} to model the source code like plain text, and shown that these models can achieve great performance on specific downstream tasks of source code. With the development of pre-trained language models in NLP, many pre-trained source code models are proposed to enhance the source code representations and efficiently be scaled to any downstream tasks~\cite{feng2020codebert,guo2021graphcodebert,nijkamp2022codegen}. Some of these models incorporate the inherent structure of code. For example, instead of taking the syntactic-level structure of source code like ASTs, \citet{guo2021graphcodebert} consider program data flow in the pre-training stage, which is a semantic-level structure of code that encodes the relation of “where-the-value-comes-from” between variables. In this survey, we focus DA methods designed for all the deep-learning-based source code models.

\subsection{What is data augmentation?} Data augmentation (DA) techniques aim 
 to improve the model's performance in terms of various aspects (e.g., accuracy and robustness) via increasing training example diversity with data synthesis. Besides, DA techniques can help avoid model overfitting in the training stage, which maintains the generability of the model. In CV, DA techniques with predefined rules are commonly adopted when training models, such as image cropping, image flipping, and color jittering~\cite{Shorten2019ASO}. These techniques can be classified as \textit{rule-based} DA. Furthermore, some attempts like Mixup have been made to create new examples by fusing multiple examples together, which is categorized as \textit{example interpolation} DA. Compared to CV, DA techniques for NLP greatly rely on language models that can help paraphrase the given context by word replacing or sentence rewriting~\cite{feng2021survey}. As most of these language models are pre-trained and can capture the semantics of inputs, they serve as reasonable frameworks to modify or paraphrase the plain text. We denote such DA methods as \textit{model-based} DA. 
 
\subsection{How does data augmentation work in source code?} Compared to images and plain texts, source code is less flexible to be augmented due to the nature of strict programming syntactic rules. Hence, we observe that most DA approaches in source code must follow the predetermined transformation rules in order to preserve the functionality and syntax of the original code snippets. To enable the complex processing of the given source code, a common approach is to use a parser to build a concrete syntax tree from the code, which represents the program grammar in a tree-like form. The concrete syntax tree will be further transformed into an abstract syntax tree (AST) to simplify the representation but maintain the key information such as identifiers, if-else statements, and loop conditions. The parsed information is utilized as the basis of the \textit{rule-based} DA approaches for identifier replacement and statement rewrite~\cite{quiring2019misleading}. From a software engineering perspective, these DA approaches can emulate more diverse code representation in real-world scenarios and thus make source code models more robust by training with the augmented data~\cite{yefet2020adversarial}.

\section{Source Code Data Augmentation Methods for Deep Learning}
\label{sec:da}
This section categorizes the mainstream DA techniques specifically designed for source code models into three parts: rule-based, model-based, and example-interpolation techniques. We explain studies of different branches as follows.

\subsection{Rule-based Techniques}
\label{subsec:rule}
\begin{figure}
    \centering
    \includegraphics[width=\columnwidth]{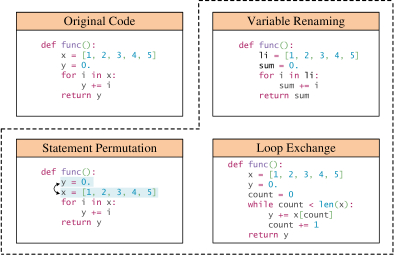}
    \caption{Rule-based DA to transform code snippets, \citet{Wang2022TestDrivenML}.}
    \label{fig:rule_da_example}
\end{figure}

A large number of DA methods utilize \textit{predetermined rules} to transform the programs without breaking syntax rules and semantics.
Specifically, these rules mainly implicitly leverage ASTs to transform the code snippets. 
The transformations can include operations such as replacing variable names, renaming method names, and inserting dead code. Besides the basic program syntax, some code transformations consider deeper structural information, such as control-flow graph (CGF) and use-define chains (UDG)~\cite{quiring2019misleading}. Additionally, a small part of rule-based DA techniques focuses on augmenting the natural language context in the code snippets, including doc-strings and comments~\cite{bahrami2021augmentedcode, song2022not, Park2023ContrastiveLW}. We illustrate a rule-based DA example relying on program grammars in Figure~\ref{fig:rule_da_example}. 

\citet{zhang2020generating} propose \texttt{MHM}, a method of iteratively renaming identifiers in the code snippets. Considered as the approach to generate examples for adversarial training, \texttt{MHM} greatly improves the robustness of source code models. Later, \citet{Srikant2021GeneratingAC} consider program obfuscations as adversarial perturbations, where they rename program variables in an attempt to hide the program’s intent from a reader. By applying these perturbed examples to the training stage, the source code models become more robust to the adversarial attack. Instead of just renaming identifiers, \texttt{BUGLAB-Aug}~\cite{allamanis2021self} contains more rules to augment code snippets, emphasizing both the programming language and natural language, such as comment deletion, comparison expression mirroring, and if-else branch swapping. The evaluation on \texttt{BUGLAB-Aug} demonstrates that DA methods can be exploited for self-supervised bug detection and repair. Similarly, \citet{jain2021contrastive} use compiler transforms as data augmentation, called \texttt{Transpiler}, automatically generating a dataset of equivalent functions. Specifically, they define 11 compiler transforms by exploiting ASTs of the programs. Rule-based DA later has been widely used for source code models to capture code representation effectively via contrastive learning~\cite{Ding2021TowardsL,liu2023contrabert}.

\citet{brockschmidt2018generative} present a generative source code model by augmenting the given AST with additional edges to learn diverse code expressions. Instead of the direct augmentation on AST, \citet{quiring2019misleading} propose three different augmentation schemes via the combination of AST and CFG, UDG and declaration-reference mapping (DRM), named as \texttt{Control Transformations}, \texttt{Declaration Transformations} and \texttt{API Transformations}. \texttt{Control Transformations} rewrite control-flow statements or modify the control flow between functions. In total, the family contains 5 transformations. This transformation involves passing variables as function arguments, updating their values, and changing the control flow of the caller and callee. \texttt{Declaration Transformations} consist of 14 transformers that modify, add or remove declarations in source code. \texttt{Declaration Transformations} make DA necessary to update all usages of variables which can be elegantly carried out using the DRM representation. \texttt{API Transformations} contain 9 transformations and exploits the fact that various APIs can be used to solve the same problem. Programmers are known to favor different APIs and thus tampering with API usage is an effective strategy for changing stylistic patterns. 

Another line of work is augmenting the natural language context in source code. \texttt{QRA}~\cite{huang2021cosqa} augments examples by rewriting natural language queries when performing code search and code question answering. It rewrites queries with minor rule-based modifications that share the same semantics as the original one. Specifically, it consists of three ways: randomly deleting a word, randomly switching the position of two words, and randomly copying a word. Inspired by this approach, \citet{Park2023ContrastiveLW} recently devised \texttt{KeyDAC} with an emphasis on the query keywords. \texttt{KeyDAC} augments on both natural language and programming language. For natural language query, it follows the rules in \texttt{QRA} but only modifies non-keywords. In terms of programming language augmentation, \texttt{KeyDAC} simply uses ASTs to rename program variables, similar to the aforementioned works.

\subsection{Model-based Techniques}
\label{subsec:model}
A series of DA techniques for source code target training various models to augment data. Intuitively, \citet{mi2021effectiveness} utilize Auxiliary Classifier Generative Adversarial Networks (\texttt{AC-GAN})~\cite{odena2017conditional} to generate augmented programs. Similarly, \citet{wang2023two} trained a generative
adversarial network to boost code generation and code search at the same time.  To increase the training data for code summarization, \texttt{CDA-CS}~\cite{song2022not} uses the pre-trained BERT model~\cite{Devlin2019BERTPO} to replace synonyms for non-keywords in code comments, which benefits the source code downstream tasks.

While these methods largely adapt the existing model-based DA techniques for general purposes, most DA approaches are specifically designed for source code models. \citet{li2022unleashing} introduce \texttt{IRGen}, a genetic-algorithm-based model using compiler intermediate representation (LLVM IR) to augment source code embeddings, where \texttt{IRGen}  generates a piece of source code into a range of semantically identical but syntactically distinct IR codes for improving model's contextual understanding. Several works~\cite{roziere2021leveraging, ahmad-etal-2023-summarize, silva2023mufin} have investigated the suitability of the multilingual generative source code models for unsupervised programming language translation via \texttt{Back-translation}, in the similar scope of the one for NLP~\cite{sennrich-etal-2016-improving}. However, unlike the one in NLP, \texttt{Back-translation} here is defined as translating between two programming languages via the natural language as an intermediate language. \citet{pinku2023pathways} exploit another generative source code model, Transcoder~\cite{roziere2020unsupervised}, to perform source-to-source translation for augmenting cross-language source code.

\subsection{Example Interpolation Techniques}
\label{subsec:ei}
Another category of data augmentation (DA) techniques, originated by \texttt{Mixup}~\cite{zhang2018mixup},
involves interpolating the inputs and labels of two or more actual examples. For instance, given that a binary classification task in CV and two images of a dog and a cat, respectively, these DA approaches like \texttt{Mixup} can blend these two image inputs and their corresponding labels based on a randomly selected weight. This collection of methods is also termed Mixed Sample Data Augmentation. Despite trials in the context of text classification problems, such methods are hard to be deployed in the realm of source code, as each code snippet is constrained by its unique program grammar and functionality.

\begin{figure}
    \centering
    \includegraphics[width=\columnwidth]{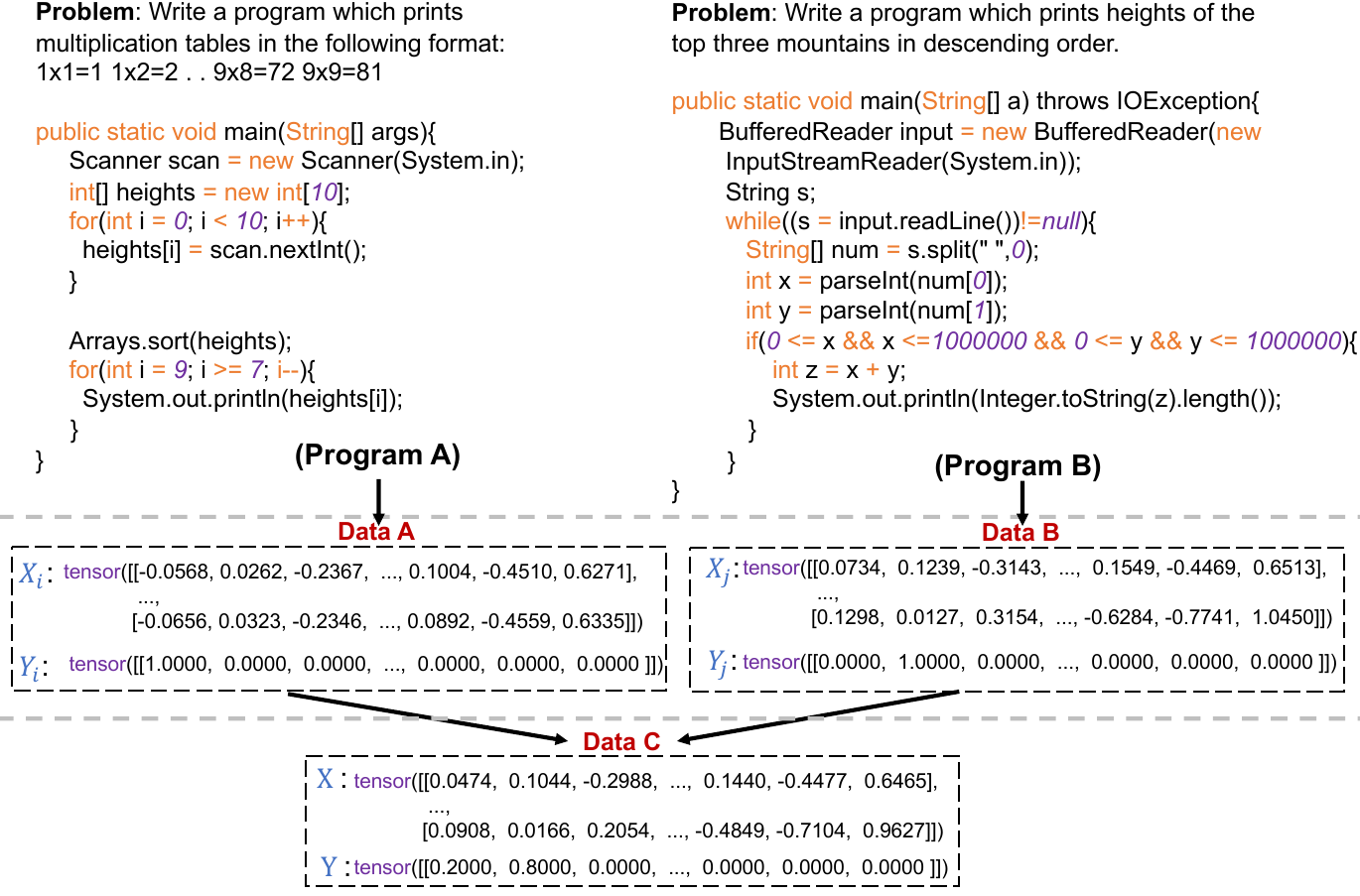}
    \caption{\texttt{MixCode}, \citet{dong2023mixcode}.}
    \label{fig:mixcode}
\end{figure}

In contrast to the aforementioned surface-level interpolation, the majority of example-interpolation DA methods are enhanced to fuse multiple real examples into a single input via model embeddings~\cite{feng2021survey}. 
As an illustration in Figure~\ref{fig:mixcode}, \citet{dong2023mixcode} merge rule-based techniques for source code models with \texttt{Mixup} to blend the representations of the original code snippet and its transformation. This approach is commonly regarded as the linear interpolation technique deployed in NLP classification tasks.

\citet{li-etal-2022-exploring-representation} introduce two novel interpolation techniques for source code models, namely \texttt{Binary Interpolation} and \texttt{Linear Extrapolation}. \texttt{Binary Interpolation} serves as a data augmentation strategy, which interchangeably swaps features between samples using elements acquired from a Bernoulli distribution. On the other hand, \texttt{Linear Extrapolation} is another data augmentation approach that generates new data points beyond the existing feature space by extending current features in accordance with a uniform distribution.

\section{Strategies and Techniques}
\label{sec:tricks}
In real-world applications, the design and efficacy of DA techniques for source code models are influenced by a variety of factors, such as computing cost, example diversity, and models' robustness. This section highlights these factors, offering insights and techniques for devising and optimizing suitable DA methods.
\subsection{Method Stacking}
\label{subsec:stack}
As discussed in Section~\ref{sec:da}, numerous DA strategies are proposed concurrently in a single work, aiming to enhance the models' performance. 
[Add one sentence to define method stacking]
Typically, the combination entails two types: same-type DA or a mixture of different DA methods. The former is typically applied in rule-based DA techniques, stemming from the realization that a single code transformation cannot fully represent the diverse code style and implementation found in the real world. 

Several works~\cite{shi2023cocosoda,huang2021cosqa} demonstrate that merging multiple types of DA techniques can enhance the performance of source code models. \citet{mi2021effectiveness} combined rule-based code transformation schemes with model-based DA using \texttt{AC-GAN} to create an augmented corpus for model training.  Instead of augmenting on programming language, \texttt{CDA-CS}~\cite{song2022not} encompasses two kinds of DA techniques: rule-based non-keyword extraction and model-based non-keyword replacement. Empirical evidence from \citet{chen2023exploring} shows that combining \texttt{Back-translation} and variable renaming can result in improved code completion performance.  

\subsection{Optimization}
\label{subsec:optimize}
In certain scenarios such as enhancing robustness and minimizing computational cost, optimally selecting specific augmented example candidates is crucial. We denote such goal-oriented candidate selections in DA as \textit{optimization}. Subsequently, we introduce three types of strategies: probabilistic, model-based, and rule-based selection. Probabilistic selection is defined as the optimization via sampling from a probability distribution, while model-based selection is guided by the model to select the most proper examples. In terms of rule-based selection, it is an optimization strategy where specific predetermined rules or heuristics are used to select the most suitable examples.

\subsubsection{Probabilistic Selection}
We introduce three representative probabilistic selection strategies, \texttt{MHM},  \texttt{QMDP}, and \texttt{BUGLAB-Aug}. \texttt{MHM}~\cite{zhang2020generating} adopts the Metropolis-Hastings probabilistic sampling method, which is a Markov Chain Monte Carlo technique, to choose adversarial examples via identifier replacement. Similarly, \texttt{QMDP}~\cite{tian2021generating} uses a Q-learning approach to strategically select and execute rule-based structural transformations on the source code, thereby guiding the generation of adversarial examples. In \texttt{BUGLAB-Aug}, \citet{allamanis2021self} model the probability of applying a specific rewrite rule at a location in a code snippet similar to the pointer net~\cite{meritypointer}.

\subsubsection{Model-based Selection}

Several DA techniques employing this strategy use the model's gradient information to guide the selection of augmented examples. An emblematic approach is the \texttt{DAMP} method~\cite{yefet2020adversarial}, which optimizes based on the model loss to select and generate adversarial examples via variable renaming. Another variant, \texttt{SPACE}~\cite{li2022semantic}, performs selection and perturbation of code identifiers' embeddings via gradient ascent, targeting to maximize the model's performance impact while upholding semantic and grammatical correctness of the programming language. A more complex technique, \texttt{ALERT}~\cite{yang2022natural}, uses a genetic algorithm in its gradient-based selection strategy. It evolves a population of candidate solutions iteratively, guided by a fitness function that calculates the model's confidence difference, aiming to identify the most potent adversarial examples.

\begin{table*}[!ht]
    \centering
    \resizebox{\linewidth}{!}{
    \begin{tabular}{l|ccccccccc} \toprule
       \textbf{DA Method} & Category & PL & NL &  Optimization &Preprocess &Parsing &Level & TA & LA\\ \midrule
       \texttt{ComputeEdge}~\cite{brockschmidt2018generative} & Rule & \cmark & \xmark & --- & --- &AST & AST & \cmark & \cmark\\
       \texttt{RefineRepresentation}~\cite{bielik2020adversarial} & Rule & \cmark & \xmark & Model & --- &AST & AST & \cmark & \cmark\\
       \texttt{Control Transformations}~\cite{quiring2019misleading}  & Rule & \cmark & \xmark & Prob & --- &AST+CFG+UDG & Input & \cmark & \xmark \\
       \texttt{Declaration Transformations}~\cite{quiring2019misleading} & Rule & \cmark & \xmark & Prob & --- &AST+DRM & Input & \cmark & \xmark\\
       \texttt{API Transformations}~\cite{quiring2019misleading} & Rule & \cmark & \xmark & Prob & --- &AST+CFG+DRM & Input & \cmark & \xmark\\
       \texttt{DAMP}~\cite{yefet2020adversarial} & Rule & \cmark & \xmark & Model & --- &AST & Input & \cmark & \cmark\\
       \texttt{IBA}~\cite{huang2021cosqa} & Rule & \xmark & \cmark & --- &Tok & --- & Embed & \xmark & \cmark\\
       \texttt{QRA}~\cite{huang2021cosqa} & Rule & \cmark & \xmark & --- &Tok & --- & Input & \xmark & \cmark\\
       \texttt{MHM}~\cite{zhang2020generating}  & Rule & \xmark & \cmark & Prob & --- &AST & Input & \cmark & \xmark\\
       \texttt{Mossad}~\cite{devore2020mossad}  & Rule & \cmark & \xmark & Rule &Tok &AST & Input & \cmark & \cmark\\
       \texttt{AugmentedCode}~\cite{bahrami2021augmentedcode} & Rule & \cmark & \xmark & --- &Tok &--- & Input & \xmark & \cmark\\
       \texttt{QMDP}~\cite{tian2021generating} & Rule & \cmark & \xmark & Prob &Tok &AST & Input & \cmark & \xmark\\
       \texttt{Transpiler}~\cite{jain2021contrastive} & Rule & \cmark & \xmark & Prob & --- &AST & Input & \cmark & \xmark\\
       \texttt{BUGLAB-Aug}~\cite{allamanis2021self} & Rule & \cmark & \xmark & Prob & Tok & AST & Input & \xmark & \cmark\\
       \texttt{SPAT}~\cite{yu2022data} & Rule & \cmark & \xmark & Model & --- & AST & Input & \cmark & \xmark \\
       \texttt{RoPGen}~\cite{li2022ropgen} & Rule & \cmark & \xmark & Model & --- & AST & Input & \cmark & \xmark \\
       \texttt{ACCENT}~\cite{zhou2022adversarial} & Rule & \cmark & \xmark & Rule & --- & AST & Input & \cmark & \cmark\\
       \texttt{SPACE}~\cite{li2022semantic} & Rule & \cmark & \xmark & Model &Tok & AST & Embed & \cmark & \cmark \\
       \texttt{ALERT}~\cite{yang2022natural} & Rule & \cmark & \xmark & Model & Tok & AST & Input & \cmark & \cmark \\
       \texttt{IRGen}~\cite{li2022unleashing} & Rule & \cmark & \xmark & Rule & --- & AST+IR & IR & \cmark & \cmark \\
       \texttt{Binary Interpolation}~\cite{li-etal-2022-exploring-representation} & EI & \cmark & \cmark  & --- & --- &--- & Embeb & \cmark & \cmark \\
       \texttt{Linear Extrapolation}~\cite{li-etal-2022-exploring-representation} & EI & \cmark & \cmark & --- & --- &--- & Embeb & \cmark & \cmark \\
       \texttt{Gaussian Scaling}~\cite{li-etal-2022-exploring-representation} & Rule & \cmark & \cmark & Model & --- &--- & Embeb & \cmark & \cmark \\
       \texttt{CodeTransformator}~\cite{zubkov2022evaluation} & Rule & \cmark & \xmark & Rule & --- & AST & Input & \cmark & \xmark \\
       \texttt{RADAR}~\cite{yang2022important} & Rule & \cmark & \xmark & Rule & --- & AST & Input & \cmark & \xmark\\
       \texttt{AC-GAN}~\cite{mi2021effectiveness} & Model & \cmark & \xmark & --- & --- & --- & Input & \cmark & \cmark \\
       \texttt{CDA-CS}~\cite{song2022not} & Model & \xmark & \cmark & Model & KWE & ---  & Input & \xmark & \cmark \\
       \texttt{srcML-embed}~\cite{li2022cross} & Rule & \cmark & \xmark & --- & --- & AST & Embed & \cmark & \xmark \\
       \texttt{MultIPA}~\cite{orvalho2022multipas} & Rule & \cmark & \xmark & --- & --- & AST & Input & \cmark & \xmark \\
        \texttt{ProgramTransformer}~\cite{rabin2022programtransformer} & Rule & \cmark & \xmark & --- & --- & AST & Input & \cmark & \xmark \\
       \texttt{Back-translation}~\cite{ahmad-etal-2023-summarize} & Model & \cmark & \xmark & --- & Tok &--- & Input & \xmark & \cmark \\
       \texttt{MixCode}~\cite{dong2023mixcode} & Rule+EI & \cmark & \cmark & --- &--- &--- & Embed & \cmark & \cmark \\
       \texttt{NP-GD}~\cite{shen4385791bash} & Model & \cmark & \xmark & Model & Tok &--- & Embed & \cmark & \cmark\\
       \texttt{ExploitGen}~\cite{yang2023exploitgen} & Rule & \xmark & \cmark & --- & --- &--- & Input & \cmark & \xmark\\
       \texttt{SoDa}~\cite{shi2023cocosoda}  & Model & \cmark & \cmark & --- & --- & AST & Input & \cmark & \cmark\\
       \texttt{Transcompiler}~\cite{pinku2023pathways} & Model & \cmark & \xmark & --- &--- &---  & Input & \cmark & \xmark \\
       \texttt{STRATA}~\cite{springer2021strata} & Rule & \cmark & \xmark &Model  & Tok  &AST & Input & \cmark & \cmark \\
       \texttt{KeyDAC}~\cite{Park2023ContrastiveLW} & Rule & \cmark & \cmark & --- &KWE &AST & Embed & \xmark & \cmark \\
       \texttt{Simplex Interpolation}~\cite{zhang2022combo} & EI & \cmark & \xmark & --- &--- &AST+IR & Embed & \xmark & \cmark \\\bottomrule
    \end{tabular}}
    \caption{Comparing a selection of DA methods by various aspects relating to their applicability, dependencies, and requirements. \textit{PL}, \textit{NL}, \textit{TA}, \textit{LA}, \textit{EI}, \textit{Prob}, \textit{Tok}, and \textit{KWE} stand for Programming Language, Natural Language, Example Interpolation, Probability, Tokenization, Keyword Extraction, Task-Agnostic, and Language-Agnostic. \textit{PL} and \textit{NL} determine if the DA method is applied to the programming language or natural language context. \textit{Preprocess} denotes preprocessing required besides the program parsing. \textit{Parsing} refers to the type of feature used by the DA method during program parsing. \textit{Level} denotes the depth at which data is modified by the DA. \textit{TA} and \textit{LA} represent whether the DA method can be applied to different tasks or programming languages. As most papers do not clearly state if their DA methods are \textit{TA} and \textit{LA}, we subjectively denote the applicability.}
    \label{tab:my_label}
\end{table*}
\subsubsection{Rule-based Selection}
Rule-based selection stands as a powerful approach, featuring predetermined fitness functions or rules. This method often relies on evaluation metrics for decision-making. For instance, \texttt{IRGen}~\cite{li2022unleashing} utilizes a Genetic-Algorithm-based optimization technique with a fitness function based on IR similarity. On the other hand, \texttt{ACCENT}~\cite{zhou2022adversarial} and \texttt{RADAR} apply evaluation metrics such as BLEU~\cite{Papineni2002BleuAM} and CodeBLEU~\cite{ren2020codebleu} respectively to guide the selection and replacement process, aiming for maximum adversarial impact. Finally,  \texttt{STRATA}~\cite{springer2021strata} employs a rule-based technique to select high-impact subtokens that significantly alter the model's interpretation of the code.

\section{Scenarios}
\label{sec:scenario}
This section delves into several commonplace scenarios of source code scenarios, where DA approaches can be applied.
\subsection{Adversarial Examples for Robustness}
\label{subsec:adv}
Robustness presents a critical and complex dimension of software engineering, necessitating the creation of semantically-preserved adversarial examples to discern and mitigate vulnerabilities within source code models. There is a surge in designing more effective DA techniques for generating these examples in recent years. Several studies~\cite{yefet2020adversarial, li2022ropgen, Srikant2021GeneratingAC, li2022semantic, anandadversarial, henke2022semantic, tian2023adversarial, yang2023assessing, li2023multi, gu2023apicom, gao2023discrete} have utilized various DA methods for testing and enhancing model robustness.  \citet{wang2022recode} have gone a step further to consolidate universally accepted code transformation rules to establish the first benchmark for source code model robustness.

\subsection{Low-Resource Domains}
\label{subsec:low}

In the domain of software engineering, the resources of programming languages are severely imbalanced~\cite{orlanski2023measuring}. While some most popular programming languages like Python and Java play major roles in the open-source repositories, many languages like Rust~\cite{wu2023rustgen} are starkly low-resource. As source code models are trained on open-source repositories and forums, the programming language resource imbalance can adversely impact their performance on the resource-scarce programming languages.
Furthermore, the application of DA methods within low-resource domains is a recurrent theme within the CV and NLP communities~\cite{Shorten2019ASO, feng2021survey}. Yet, this scenario remains underexplored within the source code discipline. 

In order to increase data in the low-resource domain for representation learning, \citet{li2022unleashing} tend to add more training data to enhance source code model embeddings by unleashing the power of compiler IR. \citet{ahmad-etal-2023-summarize} propose to use source code models to perform \texttt{Back-translation} DA, taking into consideration the scenario of low-resource programming languages.  Meanwhile, \citet{chen2023exploring} underscore the fact that source code datasets are markedly smaller than their NLP equivalents, which often encompass millions of instances. As a result, they commence investigations into code completion tasks under this context and experiment with \texttt{Back-translation} and variable renaming. \citet{shen4385791bash} contend that the generation of bash comments is hampered by a dearth of training data and thus explore model-based DA methods for this task.

\subsection{Retrieval Augmentation}
\label{subsec:retrievalaug}
Increasing interest has been observed in the application of DA for retrieval augmentation within NLP~\cite{mialon2023augmented} and source code~\cite{lu2022reacc}. These retrieval augmentation frameworks for source code models incorporate retrieval-augmented examples from the training set when pre-training or fine-tuning source code models. This form of augmentation enhances the parameter efficiency of models, as they are able to store less knowledge within their parameters and instead retrieve it. It is shown as a promising application of DA in various source code downstream tasks, such as code summarization~\cite{zhang2020retrieval,liuretrieval,yu2022bashexplainer,choi2023readsum,ye2023tram}, code completion~\cite{parvez2021retrieval,tang2023domain} and program repair~\cite{nashid2023retrieval,jin2023inferfix,wang2023rap}.

\subsection{Contrastive Learning}
\label{subsec:contrastive}
Another source code scenario to deploy DA methods is contrastive learning, where it enables models to learn an embedding space in which similar samples are close to each other while dissimilar ones are far apart~\cite{chen2022varclr,wang2022heloc,zhang2022combo, yang2023pre, ding2023concord}. As the training datasets commonly contain limited sets of positive samples, DA methods are preferred to construct similar samples as the positive ones. \citet{liu2023contrabert} make use of contrastive learning with DA to devise superior pre-training paradigms for source code models, while some works study the advantages of this application in some source code tasks like defect detection~\cite{cheng2022path}, clone detection~\cite{zubkov2022evaluation,wang2022bridging} and code search~\cite{shi2022cross,shi2023cocosoda, li2022coderetriever}.
\section{Downstream Tasks}
\label{sec:task}
In this section, we discuss several DA works for common source code tasks and evaluation datasets.

\subsection{Code Authorship Attribution}
\label{subsec:author}
Code authorship attribution is the process of identifying the author of a given code, usually achieved by source code models. \citet{yang2022natural} initially investigate generating adversarial examples on the \textit{Google Code Jam} (GCJ) dataset, which effectively fools source code models to identify the wrong author of a given code snippet. By training with these augmented examples, the model's robustness can be further improved. \citet{li2022ropgen} propose another DA method called \texttt{RoPGen} for the adversarial attack and demonstrate its efficacy on GCJ. \citet{dong2023boosting} empirically study the effectiveness of several existing DA approaches for NLP on several source code tasks, including authorship attribution on \textit{GCJ}.
\subsection{Clone Detection}
\label{subsec:clone}
Code clone detection refers to the task of identifying if the given code snippet is cloned and modified from the original sample, and can be called plagiarism detection in some cases~\cite{zubkov2022evaluation,pinku2023pathways, hasija2023neuro}. This is a challenging downstream task as it needs the source code model to understand the source code both syntactically and semantically. \citet{jain2021contrastive} propose correct-by-construction DA via compiler information to generate many variants with equivalent functionality of the training sample and show its effectiveness of improving the model robustness on  \textit{BigCloneBench}~\cite{svajlenko2014towards} and a self-collected JavaScript dataset. \citet{jia2023clawsat} show that when training with adversarial examples via obfuscation transformation, the robustness of source code models can be significantly improved. \citet{zubkov2022evaluation} provide the comparison of multiple contrastive learning frameworks, combined with rule-based transformations for the clone detection task. \citet{pinku2023pathways} later use \texttt{Transcompiler} to translate between limited source code in Python and Java and increase the training data for cross-language code clone detection.

\subsection{Defect Detection and Repair}
\label{subsec:defect}
Defect Detection and Repair, in other words, bug or vulnerability detection and fix, is to capture the bugs in given code snippets~\cite{cheng2022path,dong2023mixcode} and generate repaired versions~\cite{drain2021deepdebug, yasunaga2021break, wang2022leveraging}. The task can be considered as the binary classification task, where the labels are either true or false. \citet{allamanis2021self} implement \texttt{BUGLAB-Aug}, a DA framework of self-supervised bug detection and repair. \texttt{BUGLAB-Aug} has two sets of code transformation rules, one is a bug-inducing rewrite and the other one is rewriting as DA. Their approach boosts the performance and robustness of source code models simultaneously. \citet{cheng2022path} present a path-sensitive code embedding technique called \texttt{ContraFlow}, which uses self-supervised contrastive learning to detect defects based on value-flow paths. \texttt{ContraFlow} utilizes DA to generate contrastive value-flow representations of three datasets (namely \textit{D2A}~\cite{zheng2021d2a}, Fan~\cite{fan2020ac} and \textit{FFMPeg+Qemu}~\cite{zhou2019devign}) to learn the (dis)-similarity among programs. \citet{Ding2021TowardsL} present a novel self-supervised
model focusing on identifying (dis)similar functionalities of source code, which outperforms the state-of-the-art models on \textit{REVEAL}~\cite{Chakraborty22learning} and \textit{FFMPeg+Qemu}~\cite{zhou2019devign}. Specifically, they design code
transformation heuristics to automatically create bugged programs and similar code for augmenting pre-training data.

\subsection{Code Summarization}
\label{subsec:code_summ}
Code summarization is considered as a task that generates a comment for a piece of the source code, and is thus also named code comment generation. \cite{zhang2020training} apply \texttt{MHM} to perturb training examples and mix them with the original ones for adversarial training, which effectively improves the robustness of source code models in summarizing the adversarial code snippets. \cite{zhang2020retrieval} develop a retrieval-augmentation framework for code summarization, relying on similar code-summary pairs to generate the new summary on \textit{PCSD} and \textit{JCSD} datasets~\cite{miceli2017parallel,hu2018summarizing}. Based on this framework, \cite{liuretrieval} leverage Hybrid GNN to propose a novel retrieval-augmented code summarization method and use it during model training on the self-collected CCSD dataset. \cite{zhou2022adversarial} generate adversarial examples of a Python dataset~\cite{wan2018improving} and \textit{JSCD} to evaluate and enhance the source code model robustness.

\subsection{Code Search}
\label{subsec:search}
Code search, or code retrieval, is a text-code task that searches code snippets based on the given natural language queries. The source code models on this task need to map the semantics of the text to the source code~\cite{li-etal-2022-exploring-representation,li2023rethinking,huang2023towards,ma2023mulcs}. \citet{bahrami2021augmentedcode} increase the code search queries by augmenting the natural language context such as doc-string, code comments and commit messages. \citet{shi2022cross} use AST-focused DA to replace the function and variable names of the data in \textit{CodeSearchNet}~\cite{husain2019codesearchnet} and \textit{CoSQA}~\cite{huang2021cosqa}. \citet{shi2023cocosoda} introduce soft data augmentation (\texttt{SoDa}), without external transformation rules on code and text. With \texttt{SoDa}, the model predicts tokens based on dynamic masking or replacement when processing \textit{CodeSearchNet}. Instead of applying rule-based DA techniques, \citet{li-etal-2022-exploring-representation} manipulate the representation of the input data by interpolating examples of \textit{CodeSearchNet}.

\subsection{Code Completion}
\label{subsec:completion}
Code completion requires source code models to generate lines of code to complete given programming challenges~\cite{brockschmidt2018generative,lu2022reacc,wang2022test}. \citet{anandadversarial} suggest that source code models are vulnerable to adversarial examples which are perturbed with transformation rules. \cite{lu2022reacc} propose a retrieval-augmented code completion framework composed of the rule-based DA module to generate on \textit{PY150}~\cite{raychev2016probabilistic} and \textit{GitHub Java Corpus} datasets~\cite{allamanis2013mining}. \citet{wang2022recode} customize over 30 transformations specifically for code on docstrings, function and variable names, code syntax, and code format and benchmark generative source code models on \textit{HumanEval}~\cite{Chen2021EvaluatingLL} and \textit{MBPP}~\cite{austin2021program}. \citet{yang2022important} devise transformations on functional descriptions and signatures to attack source code models and show that their performances are susceptible.

\subsection{Code Translation}
\label{subsec:translation}
Similar to neural machine translation in NLP~\cite{stahlberg2020neural}, the task is to translate source code written in a specific programming language translation to another one. \citet{ahmad-etal-2023-summarize} apply data augmentation through back-translation to enhance unsupervised code translation. They use pre-trained sequence-to-sequence models to translate code into natural language summaries and then back into code in a different programming language, thereby creating additional synthetic training data to improve model performance. 
\citet{chen2023exploring} utilize \texttt{Back-translation} and variable augmentation techniques to yield the improvement in code translation on \textit{CodeTrans}~\cite{lu2021codexglue}.
Recently, \cite{xie2023data} have proposed two novel data augmentation methods to generate code translation pairs with similar functionality, via model training and back-translation.

\subsection{Code Question Answering (CQA)}
\label{subsec:cqa}
CQA can be formulated as a task where the source code models are required to generate a textual answer based on given a code snippet and a
question.  \citet{huang2021cosqa} incorporate two rule-base DA methods on code and text to create examples for contrastive learning. \citet{li2022semantic} explore the efficacy of adversarial training on the continuous embedding space with rule-based DA on \textit{CodeQA}~\cite{liu2021codeqa}, a free-form CQA dataset. \citet{Park2023ContrastiveLW} evaluate \texttt{KeyDAC}, a framework using query writing and variable renaming as DA, on \textit{WebQueryTest} of CodeXGLUE~\cite{lu2021codexglue}. Different from \textit{CodeQA}, \textit{WebQueryTest} is a CQA benchmark only containing Yes/No questions.

\subsection{Code Classification}
\label{subsec:problem_class}
The task performs the categorization of programs regarding their functionality~\cite{zhang2020generating,dong2023mixcode} or readability~\cite{mi2021effectiveness,mi2022enhanced,mi2022improving}. \citet{wang2022heloc} propose a novel AST hierarchy representation for contrastive learning with the graph neural network. Specifically, they augment the node embeddings in AST paths on \textit{OJ}, a dataset containing 104 classes of programs. \citet{zhang2022combo} incorporate simplex interpolation, an example-interpolation DA approach on IR, to create intermediate embeddings on \textit{POJ-104} from CodeXGLUE~\cite{lu2021codexglue}. \citet{dong2023mixcode} also explore the example-interpolation DA to fuse the embeddings of code snippets. They evaluate the method on two datasets, \textit{JAVA250} and \textit{Python800}~\cite{puri2021codenet}.

\subsection{Method Name Prediction}
\label{subsec:name_pred}
The goal of method name prediction is to predict the name of a method given the program. \citet{yefet2020adversarial} attack and defense source code models by using variable-name-replaced adversarial programs on the \textit{Code2Seq} dataset~\cite{alon2018codeseq}. \citet{ pour2021search} propose a search-based testing framework specifically for adversarial robustness. They generate adversarial examples of Java with ten popular refactoring operators widely used in Java. \citet{rabin2021generalizability} and \citet{yu2022data} both implement data augmentation frameworks and various transformation rules for processing Java source code on the \textit{Code2Seq} dataset.

\subsection{Type Prediction}
\label{subsec:type_pred}
Type prediction, or type interference, aims to predict parameter and function types in programs. \citet{bielik2020adversarial} conduct adversarial attacks on source code models with examples of transformed ASTs. They instantiate the attack to type prediction on JavaScript and TypeScript. \citet{jain2021contrastive} apply compiler transforms to generates many variants of programs in DeepTyper~\cite{hellendoorn2018deep}, with equivalent functionality with 11 rules. \citet{li2022cross} incorporate srcML~\cite{collard2013srcml} meta-grammar embeddings to augment the syntactic features of examples in three datasets, \textit{DeepTyper}, \textit{Typilus Data}~\cite{allamanis2020typilus} and \textit{CodeSearchNet}~\cite{husain2019codesearchnet}.

\section{Challenges and Opportunities}
\label{sec:challenge}
When it comes to source code, DA faces significant challenges. Nonetheless, it's crucial to acknowledge that these challenges pave the way for new possibilities and exciting opportunities in this area of work.

\paragraph{Discussion on theory.} Currently, there's a noticeable gap in the in-depth exploration and theoretical understanding of DA methods in source code. Most existing research on DA is centered around image processing and natural language fields, viewing data augmentation as a way of applying pre-existing knowledge about data or task invariance~\cite{dao2019kernel, wu2020generalization, shi2022improving}. When shifting to source code, much of the previous work introduces new methods or demonstrates how DA techniques can be effective for subsequent tasks. However, these studies often overlook why and how particularly from a mathematical perspective. With source code being discrete by nature, having a theoretical discussion becomes even more important. It allows us to understand DA from a broader perspective, not just by looking at experimental results. By exploring DA in this way, we can better understand its underlying principles without being solely dependent on experimental validation.

\paragraph{More study on pre-trained models.}
In recent years, pre-trained source code models have been widely applied in source code, containing rich
knowledge through self-supervision on a huge scale of corpora~\cite{feng2020codebert,guo2021graphcodebert, zhuo2023large}. Numerous studies have been conducted utilizing pre-trained source code models for the purpose of DA, yet, most of these attempts are confined to mask token replacement~\cite{shi2023cocosoda}, direct generation after fine-tuning~\cite{ahmad-etal-2023-summarize,pinku2023pathways}. An emergent research opportunity lies in exploring the potential of DA in the source code domain with the help of large language models (LLMs) trained on a large amount of text and source code~\cite{Chen2021EvaluatingLL, Li2023StarCoderMT}. LLMs have the capability of context generation based on prompted instructions and provided examples, making them a choice to automate the DA process in NLP~\cite{yoo2021gpt3mix,wang2021want}. Different from the previous usages of pre-trained models in DA, these works open the era of ``prompt-based DA''. In contrast, the exploration of prompt-based DA in source code domains remains a relatively untouched research area. Another direction is to harness the internal knowledge encoded in pre-trained source code models. For example, \citet{karmakar2021pre, wan2022they} show that ASTs and code semantics can be induced from these models without the static analysis tools. As most DA methods for source code models tend to predefine the code transformation rules via program analysis, it is expected that the programming knowledge inside these pre-trained source code models can automate the rule designs.

\paragraph{Working with domain-specific data.} Our paper focuses on surveying DA techniques for common downstream tasks involving processing source code. However, we are aware that there are a few works on other task-specific data in the field of source code. For instance, API recommendation and API sequence generation can be considered a part of source code tasks~\cite{huang2018api,gu2016deep}. DA methods covered by our survey can not be directly generalized to these tasks, as most of them only target program-level augmentation but not API-level. We observe a gap of DA techniques between these two different layers~\cite{treude2016augmenting,xu2020incorporating,wang2021plot2api}, which provides opportunities for future works to explore. Additionally, the source code modeling has not fully justified DA for out-of-distribution generalization. Previous studies~\cite{hajipour2022simscood, hu2022codes} assume the domain as the programs with different complexity, syntax, and semantics. We argue that this definition is not natural enough. Similar to the subdomains in NLP, like biomedical and financial texts, the application subdomains of source code can be diverse. For example, the programs to solve data science problems can significantly differ from those for web design. We encourage SE and ML communities to study the benefits of DA when applied to various application subdomains of source code.

\paragraph{More exploration on project-level source code and low-resource programming languages.} The existing methods have made sufficient progress in function-level code snippets and common programming languages. The emphasis on code snippets at the function level fails to capture the intricacies and complexities of programming in real-world scenarios, where developers often work with multiple files and folders simultaneously. Therefore, we highlight the importance of exploring DA approaches on the project level. The DA on source code projects can be distinct from the function-level DA, as it may involve more information such as the interdependencies between different code modules, high-level architectural considerations, and the often intricate relationship between data structures and algorithms used across the project~\cite{mockus2002two}. At the same time, limited by data resources~\cite{husain2019codesearchnet,orlanski2023measuring}, augmentation methods
of low-resource languages are scarce, although they have more demand for DA. Exploration in these two directions is still limited, and they could be promising directions.

\paragraph{Mitigating social bias.} As source code models have advanced software development, they may be used to develop human-centric applications such as human resources and education, where biased programs may result in unjustified and unethical decisions for underrepresented people~\cite{zhuo2023exploring}. While social bias in NLP has been well studied and can be mitigated with DA~\cite{feng2021survey}, the social bias in source code has not been brought to attention. For example, \citet{zhuo2023exploring} and \citet{liu2023uncovering} find that LLMs of source code have server bias in various demographics such as gender, sexuality, and occupation when performing code generation based on the natural language queries. To make these models more responsible in source code, we urge more research on mitigating bias. As prior works in NLP suggested, DA may be an effective technique to make source code models more responsible.

\paragraph{Few-shot learning.}
In few-shot scenarios, models are required to achieve performance that rivals that of traditional machine learning models, yet the amount of training data is extremely limited. DA methods provide a direct solution to the problem. However, limited works in few-shot scenarios
have adopted DA methods~\cite{nashid2023retrieval}. Mainstream
pre-trained source code models obtain rich semantic knowledge through language modeling. Such knowledge even covers, to some extent, the semantic information introduced by traditional paraphrasing-based DA methods. In other words, the improvement space that traditional DA methods bring to pre-trained source code models has been greatly compressed. Therefore, it is an interesting question how to provide models with fast generalization and problem-solving capability by generating high-quality augmented data in few-shot scenarios.

\paragraph{Multimodal applications.}
It is important to note that the emphasis on function-level code snippets does not accurately represent the intricacies and complexities of real-world programming situations. In such scenarios, developers often work with multiple files and folders simultaneously.s have also been developed. \citet{wang2021plot2api} and \citet{liu2022matcha} explore the chart derendering with an emphasis on source code and corresponding APIs. \citet{suris2023vipergpt} propose a framework to generate Python programs to solve complex visual tasks including images and videos. Although such multimodal applications are more and more popular, no study has yet been conducted on applying DA methods to them. A potential challenge for the multimodal source code task technique is to effectively bridge between the embedding representations for each modality in source code models, which has been investigated in vision-language multimodal tasks~\cite{ray2019sunny, tang2020semantic, hao2023mixgen}.

\paragraph{Lack of unification.} The current body of literature on data augmentation (DA) for source code presents a challenging landscape, with the most popular methods often being portrayed in a supplementary manner. A handful of empirical studies have sought to compare DA methods for source code models~\cite{de2023detecting,dong2023boosting}. However, none of these works leverages most of the existing advanced DA methods for source code models. Whereas there are well-accepted
frameworks for DA for CV (e.g. default augmentation libraries in PyTorch, RandAugment~\cite{cubuk2020randaugment}) and DA for NLP (e.g. NL-Augmenter~\cite{dhole2021nlaugmenter}), a corresponding library of generalized DA techniques for source code models is conspicuously absent. Furthermore, as existent DA methods are usually evaluated with various datasets, it is hard to determine the efficacy. Therefore, we posit that the progression of DA research would be significantly facilitated by the establishment of standardized and unified benchmark tasks, along with datasets, for the purpose of contrasting and evaluating the effectiveness of different augmentation methods. This would pave the way towards a more systematic and comparative understanding of the benefits and limitations of these methods.

\section{Conclusion}
Our paper comprehensively analyzes data augmentation techniques in the context of source code. We first explain the concept of data augmentation and its function. We then examine the primary data augmentation methods commonly employed in source code research and explore augmentation approaches for typical source code applications and tasks. Finally, we conclude by outlining the current challenges in the field and suggesting potential directions for future source code research. In presenting this paper, we aim to assist source code researchers in selecting appropriate data augmentation techniques and encourage further exploration and advancement in this field.

\section*{Limitations}
While the work presents in this paper has its merits, we acknowledge the several limitations. Firstly, our work only surveys imperative programming languages used for general-purpose programming. Therefore, some DA methods for declarative languages~\cite{zhuo2023robustness} or minor downstream tasks like cryptography misuse detection~\cite{de2023detecting}, including SQL. Secondly, our focus has been primarily on function-level DA within the source code context. As such, there remains a need for future development in project-level DA methods. Nonetheless, this paper offers a valuable collection of general-purpose DA techniques for source code models, and we hope that it can serve as an inspiration for further research in this area. Thirdly, given the page limits, the descriptions presented in this survey are essentially brief in nature. Our approach has been to offer the works in meaningful structured groups rather than unstructured sequences, to ensure comprehensive coverage. This work can be used as an index where more detailed information can be found in the corresponding works. Lastly, it is worth noting that this survey is purely qualitative and does not include any experiments or empirical results. To provide more meaningful guidance, it would be helpful to conduct comparative experiments across different DA strategies. We leave this as a suggestion for future work.
\bibliographystyle{acl_natbib}
\bibliography{custom}

\end{document}